# A Continuous Encoding-Based Representation for Efficient Multi-Fidelity Multi-Objective Neural Architecture Search

Zhao Wei[1, 2], Chin Chun Ooi[1, 2, *], Yew Soon Ong[2, 3]

[1] Institute of High Performance Computing (IHPC), Agency for Science, Technology and Research (A*STAR), Singapore
[2] Centre for Frontier AI Research (CFAR), Agency for Science, Technology and Research (A*STAR), Singapore
[3] College of Computing and Data Science (CCDS), Nanyang Technological University (NTU), Singapore

*Abstract*—Neural architecture search (NAS) is an attractive approach to automate the design of optimized architectures but is constrained by high computational budget, especially when optimizing for multiple, important conflicting objectives. To address this, an adaptive Co-Kriging-assisted multi-fidelity multi-objective NAS algorithm is proposed to further reduce the computational cost of NAS by incorporating a clustering-based local multi-fidelity infill sampling strategy, enabling efficient exploration of the search space for faster convergence. This algorithm is further accelerated by the use of a novel continuous encoding method to represent the connections of nodes in each cell within a generalized cell-based U-Net backbone, thereby decreasing the search dimension (number of variables). Results indicate that the proposed NAS algorithm outperforms previously published state-of-the-art methods under limited computational budget on three numerical benchmarks, a 2D Darcy flow regression problem and a CHASE_DB1 biomedical image segmentation problem. The proposed method is subsequently used to create a wind velocity regression model with application in urban modelling, with the found model able to achieve good prediction with less computational complexity. Further analysis revealed that the NAS algorithm independently identified principles undergirding superior U-Net architectures in other literature, such as the importance of allowing each cell to incorporate information from prior cells.

## 1. Introduction

Deep convolutional neural networks (CNNs) have been successfully applied to various domains such as image classification and segmentation. A key contributor to this success is the promulgation of various classes of neural architectures (typically proposed by experienced academic and industrial experts), which have allowed creation of fairly high-performing models even with limited computational resources [1-4]. However, the adaptation of these architectures to specific applications can still be computationally expensive, with a need for considerable manpower and material resources [5]. The design of neural architectures is dependent on specific domain knowledge and experience and optimized parameters for one problem may not transfer to other applications directly, thereby necessitating repeated optimizations for various tasks and considerable manual effort [6].

To address the above problems, neural architecture search (NAS) [7] has been proposed as an approach to the design of deep neural networks, especially in the context of automated machine learning (AutoML). In NAS, architectures are determined based on optimizing a generic backbone network using different kinds of parameterized cells, blocks, and connections. This allows for discovery of performant architectures in a wide search space (albeit pre-defined) with respect to specific objectives [8]. Nonetheless, the NAS process involves repeated, computationally expensive model training, and efficient algorithms are critical for NAS to become truly practical and wide-spread [9].

Current research on NAS methods is mainly divided into three categories, i.e., reinforcement learning (RL)-based NAS [10], gradient descent (GD)-based NAS [11], and evolutionary algorithm (EA)-based NAS [12]. RL-based NAS utilizes a reward mechanism to guide the search towards an optimum through an iterative process, but the search cost can be exorbitant [13]. GD-based NAS is more computationally efficient by virtue of gradient descent techniques, although the GPU memory required for building a supernet can be prohibitive [13]. In addition, recent RL-based and GD-based NAS methods are still not easily extended to multi-objective NAS. Both RL-based and GD-based NAS typically only focus on predictive performance while neglecting other objectives such as the network's computational complexity [14]. Generally, predictive performance has strong positive correlation with the computational complexity of the networks. Hence, both RL-based and GD-based approaches favor neural architectures with higher computational complexity, a major concern when the sustainability of deep learning is increasingly being questioned [15]. In that context, multi-objective NAS algorithms which allow for trade-offs between predictive performance and computational complexity are becoming increasingly desirable.

---

* Corresponding Author. E-mail address: ooicc@cfar.a-star.edu.sg.

EA-based methods are flexible with respect to encoding of parameters, and are well-suited to uncovering optima under multiple objectives. In particular, EA-based NAS, which is based on defining a population of network architectures and using evolutionary operations to uncover new high-quality architectures, has been cited as a viable approach for multi-objective NAS [16], with several recent work utilizing conventional evolutionary algorithms (e.g., generic algorithm (GA)) and their variants [17, 18] for NAS. Nonetheless, EA-based NAS is computationally expensive, motivating the incorporation of surrogates as a means of reducing the computational budgets [19, 20]. Despite the use of surrogates, the wide-spread use of discrete encoding (e.g., for connections between nodes), while intuitive and straightforward, results in a large combinatorial search space (curse of dimensionality) that is both challenging to explore efficiently, and to generate enough data for constructing performant surrogates. Hence, efficient encoding for the network architecture, which mitigates the curse of dimensionality, is essential for making surrogate-assisted EA-based NAS practical, scalable, and effective. Continuous encoding schemes such as random key and its variants [21, 22] have been proposed to reduce the search space's dimensionality in other problems, facilitating quicker exploration but have not been shown to work in a surrogate-assisted EA setting. The combination of a continuous encoding scheme with surrogate-assisted EA can potentially further enhance the efficiency and scalability of NAS by melding the advantages of both methods.

In real-world applications, models with various fidelities are frequently also used in expensive optimization problems such as NAS [23]. The high-fidelity (HF) model (e.g., model trained with full data and to convergence) is able to get accurate predictive results, albeit with an increase in computational cost. On the contrary, a low-fidelity (LF) model (e.g., model trained with reduced data or small number of epochs) may share only partial correlation with the high-fidelity model, but offers the advantage of being evaluated at a much lower computational cost. Hence, a multi-fidelity NAS framework is a practical route to further reduce the computational burden.

U-Nets [4] are a ubiquitous deep learning model architecture that has been successfully applied in multiple domains, including partial differential equation (PDE) surrogate modelling, image segmentation, and generative models (e.g., diffusion models). The key defining characteristic of U-Net lies in its symmetric encoder-decoder structure, which seamlessly combines global contextual information with local fine-grained details. While U-Net's default architecture is highly effective, its performance can be further enhanced by tailoring its design to specific tasks and datasets. A computationally efficient NAS that can effectively identify U-Net variants with superior accuracy and generalizability can greatly relieve the burden of deep learning practitioners today that primarily resort to simple, manual search processes when using U-Nets, a task of great importance given the aforementioned diverse domains in which U-Nets have shown applicability.

Hence, the purpose of this work is to solve a key challenge in NAS of improving search convergence and reducing search cost by introducing a continuous encoding representation and demonstrating its compatibility with two other key strategies, the use of models with different fidelities and surrogate models. With the recent focus on sustainability of AI models in deployment, there is an increasing need for deep learning models which can balance the opposing need for improved predictive performance (which scales with an increase in parameters) and computational complexity. Drawing inspiration from the success of U-Net and its variants [4, 24, 25] across various domains, we choose to demonstrate our proposed approach for effective and efficient NAS on a U-shape network, focusing on the two conflicting objectives of model predictive performance and computational complexity, a canonical example of a multi-objective scenario where the computational budget of NAS is further exacerbated. The key contributions of this work are outlined as below:

1) An improved U-Net representation that facilitates integration with NAS is developed by stacking the parameterized downsampling cells and upsampling cells with different convolution operators and connections. This includes a novel continuous encoding method that can efficiently describe the connections with corresponding operators in each cell, thereby allowing a reduction in the number of variables and more effective exploration of the reduced search space.
2) An adaptive Co-Kriging-assisted multi-fidelity multi-objective NAS algorithm is proposed to address the generalized U-Net-based NAS problem, aiming to further accelerate the search process and decrease computational costs. The proposed NAS algorithm is empirically demonstrated to work with the continuous encoding scheme described in 1).
3) Empirically, we find that the two strategies described in 1) and 2) result in U-Net variants that resemble the ResNet [1] framework, a key innovation in prior literature that was shown to greatly improve performance by ensuring efficient information flow from previous cells (layers). This design outperforms the typical generalized U-Net structure, with the final NAS-derived model achieving superior performance and efficiency compared with the state-of-the-art models reported in recent literature.

The rest of this paper is structured as follows. Section 2 reviews related work to the proposed method. Section 3 presents the generalized U-Net-based NAS problem including formulation, search space, and continuous encoding method. Section 4 outlines the proposed NAS algorithm. Sections 5 and 6 present the experimental results and analysis. Conclusions are summarized in Section 7.

## 2. Related Work

This section provides a summary of related research on EA-based NAS, multi-fidelity NAS and NAS encoding.

*2.1. Evolutionary Algorithm-Based Neural Architecture Search*

The EA-based NAS employs population-based heuristic evolutionary algorithms to identify the optimal architecture, with each individual in the population representing a candidate architecture for evolution. In particular, EA-based NAS has played an important role for automated CNN architecture design due to its flexibility in encoding and ability to find a global optimum.

A lot of researchers conduct NAS based on conventional evolutionary algorithms such as GA and particle swarm optimization (PSO). For example, Real et al. [26] introduced a large-scale evolutionary NAS method with intuitive mutation operators for image classification. Xue et al. [27] similarly proposed a multi-objective GA for NAS that balances precision and time, achieving 73.6% classification accuracy on the ImageNet dataset. Huang et al. [6] proposed a variable-length PSO method with two levels for compact NAS based on both micro and macro search spaces. Zhang et al. [28] investigated a reinforced I-Ching divination evolutionary algorithm which also uses a variable-architecture encoding strategy to search across different layers, channels, and connections. However, the search cost of these conventional EA-based NAS remains extremely high, impeding the ability to run NAS till convergence.

To solve this problem, surrogate-assisted EAs have been applied to NAS to further decrease the computational cost by replacing the time-consuming model evaluation process with surrogates. For instance, Tian et al. [29] developed a surrogate-assisted evaluator with the continuous variable-length encoding approach to facilitate NAS on image classification tasks. Fan and Wang [30] proposed a network embedding method to improve the predictive performance of surrogates during NAS, with an infill criterion that considers both convergence and uncertainty for surrogate refinement. Zhou et al. [31] proposed a surrogate-assisted cooperative co-evolutionary method with the purpose of improving the search efficiency for the generative liquid state machine, where a random forest model serves as the surrogate performance estimator. Liu et al. [32] proposed a surrogate evolutionary graph neural architecture search algorithm where multiple surrogates are applied for the fitness prediction. More related applications can be found in [33-35]. While promising, the surrogate-assisted EA-based NAS can still be computationally expensive, necessitating further development of surrogate model management strategies.

*2.2. Multi-Fidelity Neural Architecture Search*

In view of the high computational cost of NAS, previous studies have explored the use of heuristics and even training-free proxies for model performance prediction. Similarly, we can consider the training of models with various parameters such as reduced number of training data, training epochs, and/or input image resolution as lower-fidelity proxies that can be used for NAS. The low-fidelity model can be evaluated fast, albeit with potentially larger error while the high-fidelity model is more accurate but comes with high computational cost.

Along this line, Zimmer et al. [36] developed Auto-PyTorch Tabular based on Bayesian optimization and Hyperband for NAS with high efficiency and robustness. Trofimov et al. [37] also proposed a Bayesian multi-fidelity method for NAS of image classification, whereby the low-fidelity model is evaluated through the knowledge distillation loss. Xu et al. [38] integrated zero-cost proxies into the multi-fidelity optimization framework to improve the search efficiency of NAS. Yang et al. [39] also accelerated the convergence of NAS through the use of multi-fidelity evaluations. Thus, it is promising to utilize a multi-fidelity optimization framework for NAS to improve design performance and decrease search cost under a limited computational budget.

*2.3. Neural Architecture Search Encoding*

The encoding scheme in NAS provides a structured and manipulable representation of neural network architectures, thereby allowing NAS algorithms to efficiently explore the search space of potential designs. Discrete encoding strategies are commonly proposed due to the simplicity and ease of implementation for NAS. For instance, Zoph et al. [40] defined a new NASNet search space with discrete encoding to design the architectures based on the dataset of interest. Xie and Yuille [41] proposed a Genetic CNN, utilizing fixed-length binary strings to encode CNN architectures. Sun et al. [42] developed a variable-length gene encoding strategy to represent a more extensive range of architectures. However, this discrete encoding can lead to a vast search space, and the encoding dimensionality grows as the number of possible connections and operations increases, making NAS increasingly challenging. Continuous encoding schemes have been proposed in other domains as they provide a more scalable search space and facilitate smooth transitions between architectures, permitting effective integration with EA [43]. Hence, efficient continuous encoding strategies are potentially key to facilitating the exploration of well-performing architectures for specific tasks or datasets.

Importantly, we note that no prior work has brought together these distinct concepts into a single continuous encoding-based, multi-fidelity, surrogate-assisted neural architecture search framework, let alone demonstrated its effectiveness on a multi-objective problem of real-world significance.

## 3. Generalized U-Net-Based Neural Architecture Search

In this section, the base NAS problem is formulated. Then, the proposed generalized U-Net-based search space, including the use of a novel continuous encoding method to reduce the dimension of this search space, are described in detail.

*3.1. Formulation of Neural Architecture Search*

Depending on the application and deployment environment, the final, "ideal" deep learning model may balance objectives such as predictive performance and computational complexity differently. For the former, mean squared error and classification accuracy are conventional evaluation metrics for regression and classification problems. For computational complexity, the inference time, number of parameters, and number of floating-point operations (FLOPs) can be selected. However, inference time can vary with *a priori* undetermined computing environment and temperature, and the number of parameters may not scale with computational complexity [12]. Thus, we choose the number of FLOPs as a straightforward metric to assess the computational complexity in this paper. The multi-objective NAS problem is formulated in Eq. (1)

$$\min \ F(x) = \{f_p(x; w^*(x)), f_c(x)\}$$
$$\text{s.t.} \ \ w^*(x) \in \text{argmin} \ \mathcal{L}(w; x) \tag{1}$$

where $x$ represents the design variable vector of the NAS problem, such that each $x$ indicates one potential CNN architecture; $w$ represents the weight vector of CNN; $F$ denotes the multi-objective vector comprising the predictive performance objective on validation dataset $f_p$ and the computational complexity objective $f_c$ (i.e., number of FLOPs in this paper); and $\mathcal{L}$ is the CNN's training loss.

As demonstrated in Eq. (1), the multi-objective NAS problem includes two levels of optimization. The lower-level optimization focuses on identifying the optimal weights for a given CNN architecture, which also corresponds to a specific value of $f_p$ and $f_c$. In contrast, the upper-level optimization searches for the best architecture and provide inputs $x$ to the lower-level optimization. The NAS model is detailed in Section 3.2 and Section 3.3. To efficiently solve the multi-objective NAS problem, a surrogate-assisted evolutionary algorithm is proposed as detailed in Section 4.

*3.2. Search Space*

U-Net [4] has proved to be a powerful base deep learning model architecture for numerous applications, including in biomedical image segmentation, fluid dynamic regression, and state-of-the-art generative models like the diffusion model. While the base structure can be similar, various permutations of the U-Net architecture are necessary to achieve optimal performance across different problems, and a NAS framework that can be flexibly applied to uncover the optimal U-Net architecture for any new problem is of practical relevance. Thus, a U-Net-type model is employed as the backbone to demonstrate the effectiveness of the proposed NAS framework. In a typical U-Net, the model comprises of both downsampling and upsampling cells connected in a U-type manner. The typical downsampling cell comprises of two 3×3 convolution operators, two rectified linear unit (ReLU) operators, and a 2×2 max pooling operator which is repeated to double the number of feature channels. Similarly, the upsampling cell comprises of two 3×3 convolution operators with two ReLU operators and a 2×2 up-convolution operator which is used for halving the feature channels. The outputs of each upsampling cell are typically concatenated with skip connection outputs from the corresponding downsampling cell of the same resolution. Based on this typical set-up, the operators in these downsampling cells (i.e., 3×3 convolution, ReLU, and 2×2 max pooling) and upsampling cells (i.e., 3×3 convolution, ReLU, and 2×2 up-convolution) and their associated hyper-parameters are further modified to improve the model predictive performance.

To enable incorporation of the structure of each cell into a NAS framework, a directed acyclic graph is utilized for the representation of the network architecture. The developed generalized U-Net structure is graphically expressed in Fig. 1. It can be seen that the generalized U-Net consists of downsampling and upsampling cells similar to the base U-Net. The main difference is that the types and execution sequences of the operators in each cell in the generalized U-Net can be changed, thereby allowing a NAS to find more optimal architectures relative to the basic U-Net model for any problem. In the downsampling and upsampling cells, each intermediate node uses add operator for two outputs from previous cells or nodes [11]. Using Downsampling Cell 3 as an example, the two inputs for Node 1 can be selected from Cell 1 or Cell 2; the two inputs for Node 2 can be selected from Cell 1, Cell 2, or Node 1; the two inputs for Node 3 can be selected from Cell 1, Cell 2, Node 1, or Node 2. Finally, the outputs of Nodes 1-3 are concatenated as the output of Downsampling Cell 3.

As shown in Fig. 1, each downsampling cell includes downsampling operators and normal operators. Downsampling operators halve the height and width of the input image while normal operators maintain the height and width of each feature map as per the inputs. Thus, if a node receives outputs from the previous downsampling cells, a downsampling operator should be selected. Otherwise, a normal operator is required to avoid inconsistencies in the feature's height and width. Similar to the downsampling cell, each upsampling cell is comprised of upsampling operators and normal operators where the upsampling operators double the height and width of the input image.

The set of downsampling, upsampling, and normal operators are selected according to their common use in CNN literature, and are listed in Table 1. The stride value is set to 2 to halve or double the height and width of feature map in the corresponding downsampling or upsampling operators. The stride is set to 1 otherwise.

For each operator in Table 1, a fixed operation sequence (i.e., operator + Batch Normalization + ReLU) is used. The choice of operators in the downsampling cells, upsampling cells and intermediate nodes defines a new architecture which is then used for model training and prediction.

Table 1 Downsampling, upsampling, and normal operators.

| Type | Operator |
|---|---|
| Downsampling operator (stride = 2) | 3×3 squeeze-and-excitation, 3×3 dilated convolution, 3×3 depthwise-separable convolution, 3×3 convolution, 2×2 average pooling, 2×2 max pooling |
| Upsampling operator (stride = 2) | 3×3 transposed squeeze-and-excitation, 3×3 transposed dilated convolution, 3×3 transposed depthwise-separable convolution, 3×3 transposed convolution |
| Normal operator (stride = 1) | Identity, 3×3 squeeze-and-excitation, 3×3 dilated convolution, 3×3 depthwise-separable convolution, 3×3 convolution |

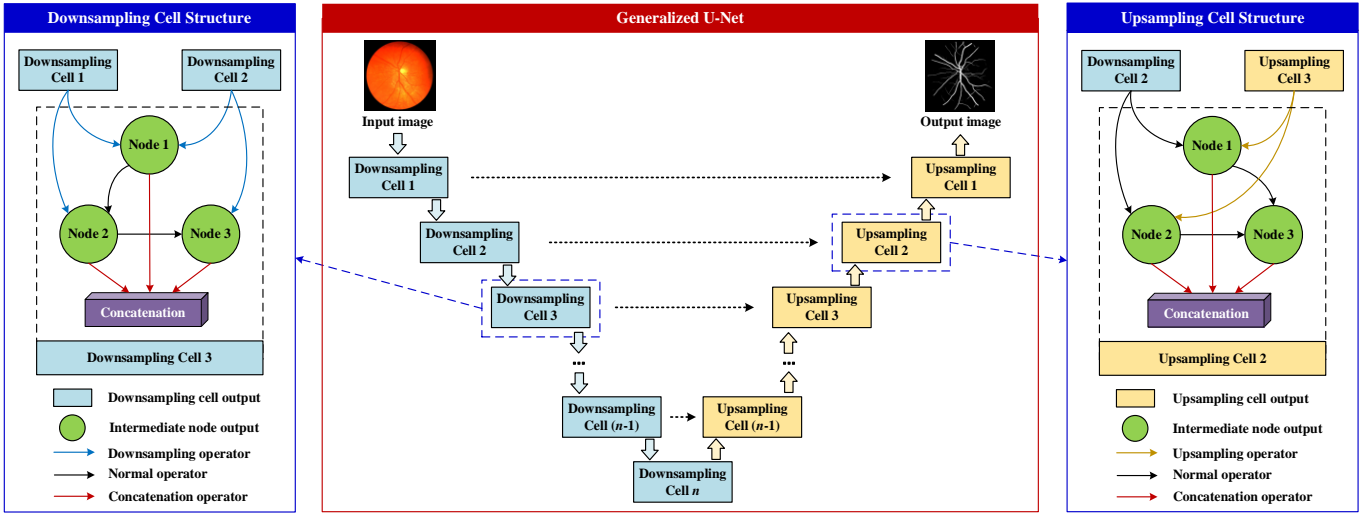

Fig. 1 Structure of the generalized U-Net.

### 3.3. Encoding

To represent the type of operator and link sequence, Ref. [40] proposed a discrete encoding method for different cells. Each intermediate node in downsampling or upsampling cells is encoded by Eq. (2) [44, 45]

$$\boldsymbol{x}_{d\_i} = \left[n_{n1\_i}, n_{o1\_i}, n_{n2\_i}, n_{o2\_i}\right], \quad i = 3, 4, \cdots, (n_n + 2) \tag{2}$$

where $\boldsymbol{x}_{d\_i}$ is the discrete encoding sequence of each node; $n_n$ represents the number of nodes; $n_{n1\_i}$ and $n_{n2\_i}$ denote the indices of the first and the second cell or node linked to Node $i$ respectively; and $n_{o1\_i}$ and $n_{o2\_i}$ represent the indices of the first and the second downsampling, upsampling or normal operators for Node $i$ respectively.

The total number of design variables is calculated by $n_c \times n_n \times 4$, where $n_c$ is the number of downsampling and upsampling cells and $n_n$ is the number of intermediate nodes in each cell. It can be seen that if $n_c$ and $n_n$ are large, the total number of design variables for NAS also grows, which is inefficient for optimization (curse of dimensionality). To effectively decrease the dimensionality of NAS, an improved continuous encoding method is proposed as in Eq. (3)

$$\boldsymbol{x}_{c\_i} = \left[n_{no1\_i}, n_{no2\_i}\right], \quad i = 3, 4, \cdots, (n_n + 2) \tag{3}$$

where $\boldsymbol{x}_{c\_i}$ is the proposed continuous encoding sequence of each node; $n_{no1\_i}$ and $n_{no2\_i}$ are the continuous encoding elements linked to the first and second cell or node for Node $i$ respectively. The definition of integer parts of $n_{no1\_i}$ and $n_{no2\_i}$ is same as $n_{n1\_i}$ and $n_{n2\_i}$, i.e., the indices of the first and the second cell or node linked to Node $i$ respectively. The fractional parts of $n_{no1\_i}$ and $n_{no2\_i}$ indicate the first and the second downsampling, upsampling or normal operators for Node $i$. The $j$-th operator of the downsampling, upsampling or normal operators is selected if the fractional part of $n_{no1\_i}$ or $n_{no2\_i}$ is located in $\left[(j-1)/n_o, j/n_o\right]$, where $n_o$ is the number of downsampling, upsampling or normal operators.

As shown in Table 1, there are 6, 4 and 5 downsampling, upsampling and normal operators respectively. Using Downsampling Cell 3 as an example, the discrete encoding and the proposed continuous encoding are graphically expressed in Fig. 2. It is evident that the proposed continuous encoding method can halve the dimensionality for NAS compared with the discrete encoding method, which is critical to further enhance the optimization efficiency of the evolutionary algorithm.

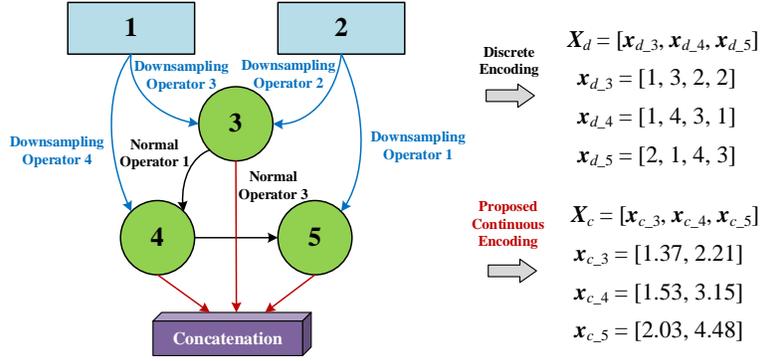

Fig. 2 Discrete encoding and proposed continuous encoding for Downsampling Cell 3.

## 4. Adaptive Co-Kriging-Assisted Multi-Fidelity Multi-Objective Differential Evolution for NAS

This section introduces an adaptive Co-Kriging-assisted multi-fidelity multi-objective differential evolution (ACK-MFMO-DE) proposed to efficiently solve the NAS problem. The conventional Co-Kriging model is briefly described first. Then, the overall procedure of ACK-MFMO-DE and its clustering-based local multi-fidelity infill sampling strategy are detailed.

*4.1. Co-Kriging Model*

A Co-Kriging model [46] fuses both HF and LF samples to provide predictions for scenarios when the HF model is computationally expensive.

To construct the Co-Kriging model, HF and LF samples, defined as $X_{HF}$ and $X_{LF}$ respectively, are used. By invoking the HF/LF models, the responses of $X_{HF}$ and $X_{LF}$ are calculated in Eq. (4)

$$Y = \begin{bmatrix} Y_{HF}(X_{HF}) \\ Y_{LF}(X_{LF}) \end{bmatrix} = \begin{bmatrix} y_{HF}(x_{HF}^{(1)}), \cdots, y_{HF}(x_{HF}^{(n_{HF})}), \\ y_{LF}(x_{LF}^{(1)}), \cdots, y_{LF}(x_{LF}^{(n_{LF})}) \end{bmatrix}^T \quad (4)$$

where $Y_{HF}(X_{HF})$ and $Y_{LF}(X_{LF})$ are the responses of $X_{HF}$ and $X_{LF}$ respectively; $n_{HF}$ and $n_{LF}$ are the number of HF and LF samples respectively.

The HF model can be approximated by scaling the LF model with a factor $\rho$ and adding the difference between the HF and LF models as expressed in Eq. (5)

$$Z_{HF}(x) = \rho Z_{LF}(x) + Z_d(x) \quad (5)$$

where $Z_{HF}$ and $Z_{LF}$ are the respective Gaussian processes describing the HF and LF models; $Z_d$ represents the difference between $Z_{HF}$ and $\rho Z_{LF}$.

Utilizing both HF/LF samples, the covariance matrix $C$ of Co-Kriging model is formulated as

$$C = \begin{bmatrix} \sigma_{LF}^2 \psi_{LF}(X_{LF}, X_{LF}) & \rho \sigma_{LF}^2 \psi_{LF}(X_{LF}, X_{HF}) \\ \rho \sigma_{LF}^2 \psi_{LF}(X_{HF}, X_{LF}) & \rho^2 \sigma_{LF}^2 \psi_{LF}(X_{HF}, X_{HF}) + \sigma_d^2 \psi_d(X_{HF}, X_{HF}) \end{bmatrix} \quad (6)$$

where $\psi_{LF}$ and $\psi_d$ are the correlation matrices of $Z_{LF}$ and $Z_d$ respectively. In this work, the Gaussian correlation function with hyperparameters $\theta = [\theta_1, \theta_2, \cdots, \theta_{n_v}]$ is shown in Eq. (7), where $n_v$ is the optimization problem dimensionality.

$$\psi(x^{(i)}, x^{(j)}) = \exp\left[-\sum_{k=1}^{n_v} \theta_k (x_k^{(i)}, x_k^{(j)})^2\right] \quad (7)$$

The $Z_{LF}$ hyperparameters $\theta_{LF}$ are determined by maximizing the logarithm of the likelihood function as shown in Eq. (8).

$$\begin{aligned} \max \quad & -\frac{n_{LF}}{2}\ln(\hat{\sigma}_{LF}^2) - \frac{1}{2}\ln\left|\det(\psi_{LF}(X_{LF}, X_{LF}))\right| \\ \hat{\mu}_{LF} &= \mathbf{1}^T \psi_{LF}(X_{LF}, X_{LF})^{-1} Y_{LF} \big/ \mathbf{1}^T \psi_{LF}(X_{LF}, X_{LF})^{-1} \mathbf{1} \\ \hat{\sigma}_{LF}^2 &= (Y_{LF} - \mathbf{1}\hat{\mu}_{LF})^T \psi_{LF}(X_{LF}, X_{LF})^{-1} (Y_{LF} - \mathbf{1}\hat{\mu}_{LF}) \big/ n_{LF} \end{aligned} \quad (8)$$

Similarly, to calculate the $Z_d$ hyperparameters $\theta_d$ and scaling factor $\rho$, the ln-likelihood function is also maximized as shown in Eq. (9).

$$\max \quad -\frac{n_{\text{HF}}}{2}\ln\left(\hat{\sigma}_d^2\right) - \frac{1}{2}\ln\left|\det\left(\psi_d\left(X_{\text{HF}}, X_{\text{HF}}\right)\right)\right|$$
$$\hat{\mu}_d = \mathbf{1}^{\text{T}}\psi_d\left(X_{\text{HF}}, X_{\text{HF}}\right)^{-1} d \big/ \mathbf{1}^{\text{T}}\psi_d\left(X_{\text{HF}}, X_{\text{HF}}\right)^{-1}\mathbf{1}$$
$$\hat{\sigma}_d^2 = \left(d - \mathbf{1}\hat{\mu}_d\right)^{\text{T}}\psi_d\left(X_{\text{HF}}, X_{\text{HF}}\right)^{-1}\left(d - \mathbf{1}\hat{\mu}_d\right)/n_{\text{HF}}$$
$$d = Y_{\text{HF}}\left(X_{\text{HF}}\right) - \rho Y_{\text{LF}}\left(X_{\text{HF}}\right) \tag{9}$$

After maximizing Eqs. (8) and (9), the hyperparameters $\hat{\theta}_{\text{LF}}$, $\hat{\theta}_d$, and $\hat{\rho}$ can be obtained. The Co-Kriging prediction of the HF model is represented as

$$\hat{y}_{\text{HF}}(x) = \hat{\mu} + c^{\text{T}}C^{-1}(Y - \mathbf{1}\hat{\mu})$$
$$c = \begin{bmatrix} \hat{\rho}\hat{\sigma}_{\text{LF}}^2\psi_{\text{LF}}(X_{\text{LF}}, x) \\ \hat{\rho}^2\hat{\sigma}_{\text{LF}}^2\psi_{\text{LF}}(X_{\text{HF}}, x) + \hat{\sigma}_d^2\psi_d(X_{\text{HF}}, x) \end{bmatrix} \tag{10}$$
$$\hat{\mu} = \mathbf{1}^{\text{T}}C^{-1}Y / \mathbf{1}^{\text{T}}C^{-1}\mathbf{1}$$

where $\hat{y}_{\text{HF}}$ is the objective prediction of the Co-Kriging model.

Note that the predicted mean squared error of the Co-Kriging model is able to be obtained through Eq. (11)
$$s^2(x) = \hat{\rho}^2\hat{\sigma}_{\text{LF}}^2 + \hat{\sigma}_d^2 - c^{\text{T}}C^{-1}c \tag{11}$$
where $s^2$ is the predicted mean squared error of the Co-Kriging model. Eq. (11) indicates that the predicted mean squared errors are zero at the HF samples. More details of the Co-Kriging model are presented in [46].

### 4.2. Overall Procedure of ACK-MFMO-DE

Generally, the U-Net training time can be quite long due to the large amount of training data and network parameters. Thus, the search cost of a NAS algorithm can be very high because the performance of each searched architecture can only be evaluated after the training has converged. To reduce the search cost and enhance the optimization efficiency, multi-fidelity models can be used for training. Models trained with the complete, full-sized data and with a large number of epochs can be treated as HF models. On the contrary, models trained with down-sampled data or a small number of epochs can be treated as LF models. Hence, an adaptive Co-Kriging-assisted multi-fidelity multi-objective differential evolution (ACK-MFMO-DE) is proposed to leverage the information from models with multiple fidelities. In this algorithm, a Co-Kriging model is established to approximate the HF/LF models, and a clustering-based local multi-fidelity infill sampling strategy is developed to adaptively update the Co-Kriging model during the optimization process, thereby allowing more efficient optimization.

The flowchart of ACK-MFMO-DE is depicted in Fig. 3, with the procedures outlined as below.

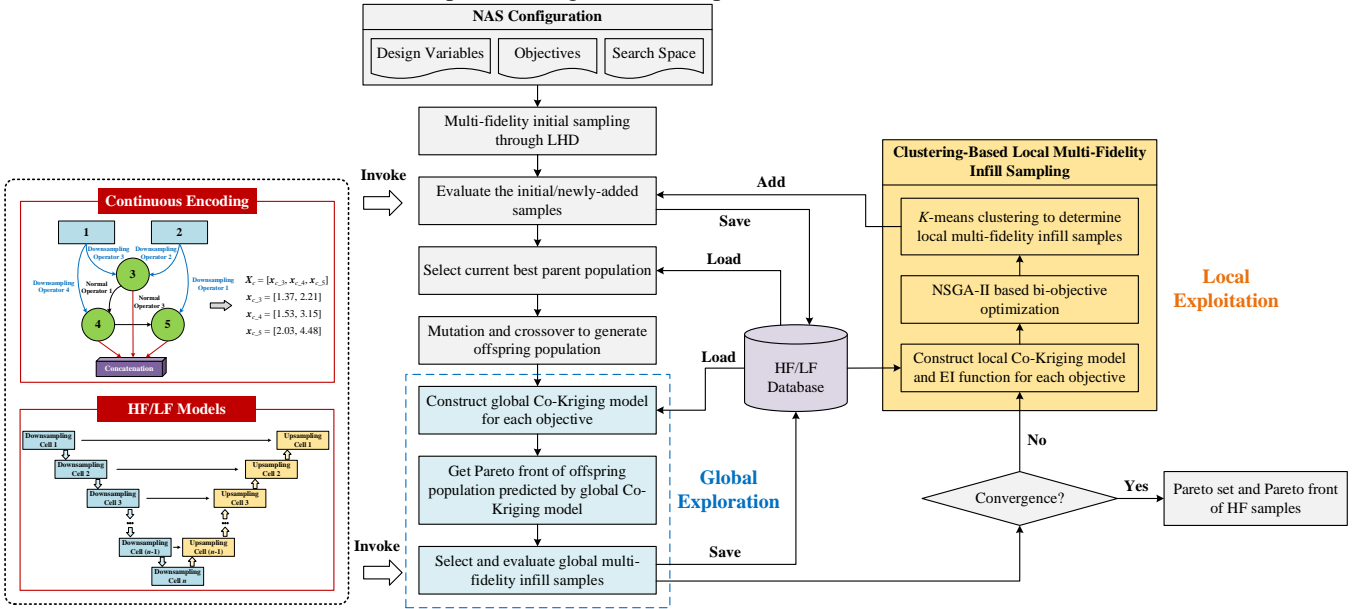

Fig. 3 Flowchart of ACK-MFMO-DE.

**Step 1:** The NAS is configured with the design variables, objectives, search space, maximum number of HF function evaluations NFE$_{\text{max\_HF}}$, initial number of HF/LF samples $n_{s\_\text{HF}}/n_{s\_\text{LF}}$, and algorithm parameters such as population size $n_p$, scaling factor $F$ and crossover rate $p_c$.

**Step 2:** The Maximin Latin Hypercube Design (LHD) is adopted for producing some initial HF/LF samples. The initial HF to LF sample ratio is set at 1:2. The HF and LF models are utilized to assess the prediction performance and computational complexity of the corresponding U-Net models.

**Step 3:** To select superior samples as parent population, a non-dominated sorting is performed for all HF samples among the database. Next, $n_p$ HF samples belonging to the outer non-dominated set of all HF samples are determined as the current parent population $X_g$.

**Step 4:** Mutation and crossover operations are conducted for generating an offspring population $U_g$. In this mutation process, several mutation operators, including DE/rand/1 (M1), DE/best/1 (M2), DE/rand/2 (M3), DE/best/2 (M4), DE/current-to-rand/1 (M5), and DE/current-to-best/1 (M6), are executed on $X_g$ to generate the mutation population $V_g = \{V_g^{(M1)} \cup V_g^{(M2)} \cup \cdots \cup V_g^{(M6)}\}$. The binomial crossover operator is used for $V_g$ to generate the offspring population $U_g = \{U_g^{(M1)} \cup U_g^{(M2)} \cup \cdots \cup U_g^{(M6)}\}$.

**Step 5:** The global Co-Kriging models are established for each objective based on all HF/LF samples in the database as shown in Eq. (12)

$$\hat{f}_{i\_\text{global}}(x) \sim \text{Co-KRG}\left(\hat{\mu}_{f_{i\_\text{global}}}, s^2_{f_{i\_\text{global}}}\right), i = 1, 2 \tag{12}$$

where $\hat{f}_{i\_\text{global}}$ is the global Co-Kriging model for the *i*-th objective; $\hat{\mu}_{f_{i\_\text{global}}}$ and $s^2_{f_{i\_\text{global}}}$ represent the predicted mean and mean squared error of the global Co-Kriging model for the *i*-th objective. In this work, the NAS problem has two objectives, i.e., prediction performance and computational complexity.

Then the established global Co-Kriging models are used for evaluating the performance of $U_g$. Through the non-dominated sorting for the predicted objectives of $U_g$, one HF sample and two other LF samples are selected randomly on the Pareto front and evaluated by the HF and LF models respectively. This step is essentially a global exploration for the NAS via the use of global Co-Kriging models, and the HF sample is the global infill sample $x_{g\_\text{global}\_HF} \in U_g$.

**Step 6:** The NAS termination criterion is evaluated. If the current number of HF function evaluations is larger than $\text{NFE}_{\max\_HF}$, the search process is halted, and the Pareto set and Pareto front of all HF samples in the database are the outputs; otherwise, the search process proceeds to Step 7.

**Step 7:** The proposed clustering-based local multi-fidelity infill sampling strategy is used to generate high-quality local HF/LF infill samples for local exploitation and refining of the Co-Kriging models, as detailed in Section 4.3. The generated local HF/LF infill samples are further evaluated before the search process returns to Step 3 for further iteration.

*4.3. Clustering-Based Local Multi-Fidelity Infill Sampling*

The clustering-based local multi-fidelity infill sampling aims to further exploit the search space around the current HF global infill sample. The pseudo-codes of the clustering-based local multi-fidelity infill sampling process are detailed in Algorithm 1, and the procedures are detailed as below.

**Step 1:** $n_{\text{near}}$ HF samples and $2n_{\text{near}}$ LF samples which are near to $x_{g\_\text{global}\_HF}$ in the database are selected to construct the local Co-Kriging models as shown in Eq. (13)

$$\hat{f}_{i\_\text{local}}(x) \sim \text{Co-KRG}\left(\hat{\mu}_{f_{i\_\text{local}}}, s^2_{f_{i\_\text{local}}}\right), i = 1, 2 \tag{13}$$

where $\hat{f}_{i\_\text{local}}$ is the local Co-Kriging model for the *i*-th objective; $\hat{\mu}_{f_{i\_\text{global}}}$ and $s^2_{f_{i\_\text{global}}}$ represent the predicted mean and mean squared error of the local Co-Kriging model for the *i*-th objective.

**Step 2:** The expected improvement (EI) function is established based on the local Co-Kriging model for each objective as formulated in Eq. (14)

$$\text{EI}_i(x) = \left(f_{i\_\min} - \hat{f}_{i\_\text{local}}(x)\right)\Phi\left(\frac{f_{i\_\min} - \hat{f}_{i\_\text{local}}(x)}{s_{f_{i\_\text{local}}}(x)}\right) + s_{f_{i\_\text{local}}}(x)\phi\left(\frac{f_{i\_\min} - \hat{f}_{i\_\text{local}}(x)}{s_{f_{i\_\text{local}}}(x)}\right), \quad i = 1, 2 \tag{14}$$

where $\text{EI}_i$ represents the EI function of the *i*-th objective; $f_{i\_\min}$ represents the minimal value of the *i*-th objective of the HF samples in the database.

**Step 3:** According to the established EI functions, the NSGA-II algorithm [47] is adopted to solve the bi-objective optimization problem as per Eq. (15). After the bi-objective optimization, the Pareto front with respect to $\text{EI}_1$ and $\text{EI}_2$ is generated.

$$\begin{aligned}\text{find} \quad & x \in [x_{\text{LB}}, x_{\text{UB}}] \\ \min \quad & \begin{cases} -\text{EI}_1(x) \\ -\text{EI}_2(x) \end{cases}\end{aligned} \tag{15}$$

**Step 4:** To further exploit the Pareto front of $\text{EI}_1$ and $\text{EI}_2$, *K*-means clustering is employed to separate the Pareto set of $\text{EI}_1$ and $\text{EI}_2$ into *K* clusters. In each cluster, the cluster center is regarded as the local HF infill sample and evaluated by the HF model. In the meantime, two other samples in each cluster are selected as local LF infill samples and evaluated by the LF model, as graphically depicted in Fig. 4.

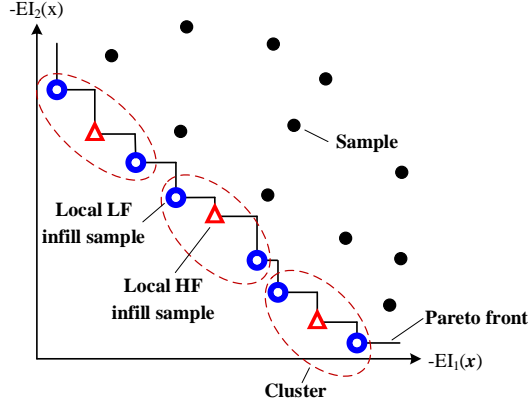

Fig. 4 Clustering-based local multi-fidelity infill sampling process.

---

**Algorithm 1** Clustering-Based Local Multi-Fidelity Infill Sampling

**Input:** HF/LF samples in the database $X_{HF}/X_{LF}$, prediction performance objective values of HF/LF samples $F_{1\_HF}/F_{1\_LF}$, computational complexity objective values of HF/LF samples $F_{2\_HF}/F_{2\_LF}$, minimal prediction performance and computational complexity objective values of HF samples $f_{1\_min}/f_{2\_min}$, number of clusters $K$.

**Output:** Local HF/LF infill samples $X_{HF\_local\_infill}/X_{LF\_local\_infill}$, prediction performance objective values of local HF/LF infill samples $F_{1\_HF\_local\_infill}/F_{1\_LF\_local\_infill}$, computational complexity objective values of local HF/LF infill samples $F_{2\_HF\_local\_infill}/F_{2\_LF\_local\_infill}$.

1    $X_{HF\_near}, F_{1\_HF\_near}, F_{2\_HF\_near}, X_{LF\_near}, F_{1\_LF\_near}, F_{2\_LF\_near} \leftarrow \text{FindNearSamples}(X_{HF}, F_{1\_HF}, F_{2\_HF}, X_{LF}, F_{1\_LF}, F_{2\_LF})$

2    $\hat{f}_{1\_local}, \hat{\mu}_{f_{1\_local}}, s^2_{f_{1\_local}}, \hat{f}_{2\_local}, \hat{\mu}_{f_{2\_local}}, s^2_{f_{2\_local}} \leftarrow \text{ConstructLocalCoKriging}(X_{HF\_near}, F_{1\_HF\_near}, F_{2\_HF\_near}, X_{LF\_near}, F_{1\_LF\_near}, F_{2\_LF\_near})$

3    $EI_1, EI_2 \leftarrow \text{ConstructEI}(\hat{f}_{1\_local}, s^2_{f_{1\_local}}, f_{1\_min}, \hat{f}_{2\_local}, s^2_{f_{2\_local}}, f_{2\_min})$

4    $X_{Pareto} \leftarrow \text{BiobjectiveOptimization}(EI_1, EI_2)$

5    $X_{HF\_local\_infill}, X_{LF\_local\_infill} \leftarrow \text{KmeansClustering}(X_{Pareto})$

6    $F_{1\_HF\_local\_infill}, F_{2\_HF\_local\_infill}, F_{1\_LF\_local\_infill}, F_{2\_LF\_local\_infill} \leftarrow \text{Evaluation}(X_{HF\_local\_infill}, X_{LF\_local\_infill})$

7    **return** $X_{HF\_local\_infill}, F_{1\_HF\_local\_infill}, F_{2\_HF\_local\_infill}, X_{LF\_local\_infill}, F_{1\_LF\_local\_infill}, F_{2\_LF\_local\_infill}$

---

## 5. Experimental Study

In this section, multiple numerical benchmarks and datasets are used to validate the proposed method.

### 5.1. Implementation Details

In this work, three multi-fidelity numerical benchmarks, one partial differential equation (PDE) dataset (i.e., 2D Darcy Flow) [48], and one medical image dataset (i.e., CHASE_DB1) [49] are utilized to validate the effectiveness of the proposed ACK-MFMO-DE and NAS model for generalized U-Net. The generalized U-Net-based NAS model as obtained by ACK-MFMO-DE is named as ACK-MFMO-DE-U-Net in this paper. To facilitate comparison, another recent, state-of-the-art multi-fidelity multi-objective algorithm MO-ACK [50] is also used to solve the NAS problem of the generalized U-Net, and the corresponding model is denoted as MO-ACK-U-Net in the subsequent results.

#### 5.1.1. Numerical Benchmarks

Three multi-fidelity variants of ZDT1, ZDT2, and ZDT3 are used to test the performance of ACK-MFMO-DE. The formulations of the multi-fidelity ZDT1, ZDT2, and ZDT3 problems are shown in Eqs. (16), (17), and (18) respectively [51, 52]

$$\begin{aligned}
&f_{1\_HF}(x) = f_{1\_LF}(x) = x_1 \\
&f_{2\_HF}(x) = g(x) \cdot h(x) \\
&f_{2\_LF}(x) = [0.8g(x) - 0.2] \cdot [1.2h(x) + 0.2] \\
&\begin{cases} g(x) = 1 + \dfrac{9}{n-1}\sum_{i=2}^{n} x_i, \ h(x) = 1 - \sqrt{x_1/g(x)} \\ 0 \le x_i \le 1, \ i = 1, 2, \cdots, n_{ZDT1} \end{cases}
\end{aligned} \tag{16}$$

$$\begin{aligned}
&f_{1\_HF}(x) = f_{1\_LF}(x) = x_1 \\
&f_{2\_HF}(x) = g(x) \cdot h(x) \\
&f_{2\_LF}(x) = [0.9g(x) + 1.1] \cdot [1.1h(x) - 0.1] \\
&\begin{cases} g(x) = 1 + \dfrac{9}{n-1}\sum_{i=2}^{n} x_i, \ h(x) = 1 - [x_1/g(x)]^2 \\ 0 \le x_i \le 1, \ i = 1, 2, \cdots, n_{ZDT2} \end{cases}
\end{aligned} \tag{17}$$

$$\begin{aligned}
&f_{1\_HF}(x) = f_{1\_LF}(x) = x_1 \\
&f_{2\_HF}(x) = g(x) \cdot h(x) \\
&f_{2\_LF}(x) = [0.75g(x) - 0.25] \cdot [1.25h(x) + 0.25] \\
&\begin{cases} g(x) = 1 + \dfrac{9}{n-1}\sum_{i=2}^{n} x_i \\ h(x) = 1 - \sqrt{x_1/g(x)} - [x_1/g(x)]\sin(10\pi x_1) \\ 0 \le x_i \le 1, \ i = 1, 2, \cdots, n_{ZDT3} \end{cases}
\end{aligned} \tag{18}$$

where $f_{1\_HF}$ and $f_{1\_LF}$ are the HF and LF models of the first objective; $f_{2\_HF}$ and $f_{2\_LF}$ are the HF and LF models of the second objective; and $n_{ZDT1} = n_{ZDT2} = n_{ZDT3} = 30$ are the respective dimensionalities of the multi-fidelity ZDT1, ZDT2, and ZDT3 problems.

5.1.2. 2D Darcy Flow Dataset

The 2D Darcy Flow is also tested as a representative regression problem for engineering systems. The steady-state solution of the 2D Darcy Flow is formulated as

$$\begin{aligned}
-\nabla(a(x)\nabla u(x)) &= f(x), \ x \in (0,1)^2 \\
u(x) &= 0, \ x \in \partial(0,1)^2
\end{aligned} \tag{19}$$

where $f$ is the force term and set to be a constant value 100; $u$ is the solution; and the viscosity term $a(x)$ can vary with different systems (input for this test problem).

The dataset of 2D Darcy Flow over the unit square is acquired from [48]. The spatial dimensionality and resolution of the 2D Darcy Flow dataset are 2 and 128×128 respectively. The number of samples in this dataset is 10000. According to [48], 90% of the dataset is used for training while the other 10% is used for testing. The Adam optimizer, along with a cosine annealing learning rate scheduler and an initial learning rate of 0.001, is used for training. The batch size is set to 32.

During NAS, models trained with 25/5 epochs are used as the HF/LF models. The training is subsequently extended to 200 epochs using the optimized architectures to get the final results. Since the 2D Darcy Flow is a regression problem, global regression metrics (i.e., root mean square error (RMSE), normalized RMSE (nRMSE), maximum error (ME)) and local regression metrics (i.e., RMSE of the conserved value (cRMSE) and RMSE at boundaries (bRMSE)) in [48] are used as the evaluation metrics. The objectives considered for the multi-objective NAS are predictive performance and computational complexity as measured by RMSE and number of FLOPs respectively.

5.1.3. CHASE_DB1 Dataset

CHASE_DB1 is a dataset for retinal vessel segmentation. It contains 28 color retina images of size 999×960 pixels, and the annotations of each image are provided by two independent human experts. In this work, 20 images are used for training and 8 images are used for testing. The Adam optimizer, combined with a cosine annealing learning rate scheduler and an initial learning rate of 0.0005, is used for training. The batch size is set to 64.

Similarly, models trained with 25/5 epochs are used as HF/LF models in the NAS process. Considering the high computational cost of training on CHASE_DB1 dataset, 20000 smaller patches are randomly extracted from each image for training during the NAS process. After searching, an additional 200000 patches are used for additional training on the optimized architectures with an increased number of epochs (i.e., 50). Five commonly used metrics, i.e., sensitivity (SE), specificity (SP), accuracy (ACC), F1 score (F1), and area under the ROC curve (AUROC), are used to judge segmentation performance. The AUROC is used as the predictive performance objective for NAS while the number of FLOPs is used to evaluate computational complexity.

The parameters of ACK-MFMO-DE for these cases are shown in Table 2.

Table 2 Parameters of ACK-MFMO-DE.

| Parameter | Value |
| --- | --- |
| Maximum number of HF function evaluations ($NFE_{max\_HF}$) | 100 |
| Initial number of HF/LF samples ($n_{s\_HF}/n_{s\_LF}$) | 50/100 |
| Population size ($n_p$) | 50 |
| Scaling factor ($F$) | 0.8 |
| Crossover rate ($p_c$) | 0.8 |

Besides, the number of downsampling and upsampling cells are set to 6 and 5 respectively, while each cell has 3 intermediate nodes. The structure of all the downsampling or upsampling cells are kept the same. Thus, the dimensionality of the NAS problem for both 2D Darcy Flow and CHASE_DB1 datasets is $2\times3+2\times3=12$ through the proposed continuous encoding method. The dimensionality of the three numerical benchmarks is 30.

All experiments are implemented in the PyTorch framework and are run on an Intel Xeon Gold 6336Y 2.40GHz CPU workstation with a NVIDIA RTX A5000 GPU. Ten runs are implemented for each algorithm on the numerical benchmarks while three runs are implemented for each algorithm on the other two NAS problems to facilitate comparison.

*5.2. Results*

5.2.1. Numerical Benchmarks

To illustrate the performance of ACK-MFMO-DE, another state-of-the-art multi-fidelity multi-objective algorithm MO-ACK [50] and three competitive surrogated-assisted evolutionary algorithms (i.e., REMO [53], DK-RVEA [54], and HSMEA [55]) are used for comparison. In each run, 100 HF samples and 200 LF samples are used for ACK-MFMO-DE and MO-ACK. Similarly, the $NFE_{max\_HF}$ for REMO, DK-RVEA, and HSMEA is set to 300 while the population size is set to 50. The iteration curves of normalized HV values with error bars are plotted in Fig. 5.

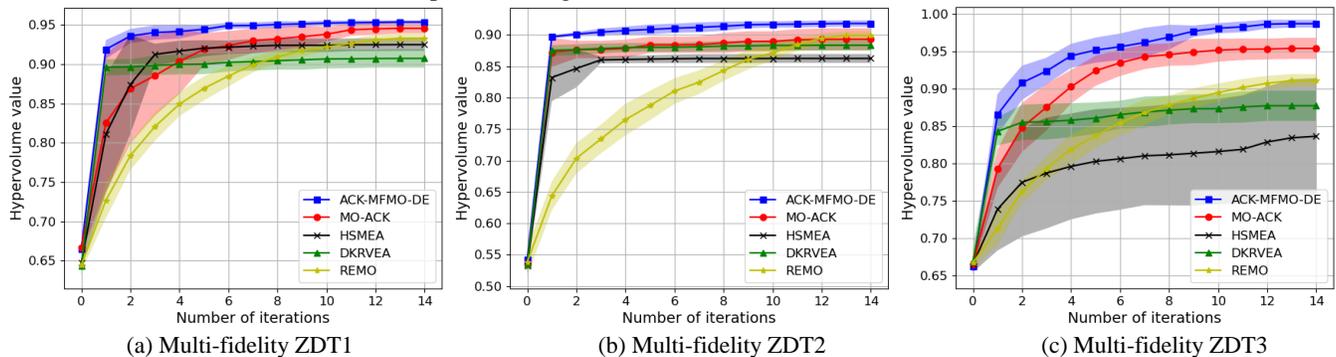

(a) Multi-fidelity ZDT1  (b) Multi-fidelity ZDT2  (c) Multi-fidelity ZDT3

Fig. 5 Iteration curves of normalized HV values with error bars for numerical benchmarks. Shaded area indicates the standard deviation on both sides of the mean.

In Fig. 5, the convergence performance of ACK-MFMO-DE outperforms MO-ACK under the same number of function evaluations. Besides, the normalized HV value obtained by ACK-MFMO-DE is significantly improved compared with that of the evolutionary algorithms which only use a high-fidelity model (i.e., REMO, DK-RVEA, and HSMEA), thereby illustrating the improved optimization performance of the proposed ACK-MFMO-DE under a limited computational budget.

5.2.2. NAS Results on 2D Darcy Flow

The results with median hypervolume (HV) value among three different runs are selected for analysis. The Pareto front of different generalized U-Net architectures obtained by ACK-MFMO-DE for 2D Darcy Flow dataset is shown in Fig. 6.

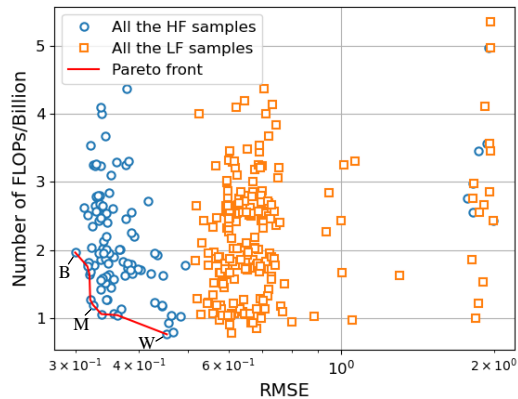

Fig. 6 Pareto front of different generalized U-Net architectures obtained by ACK-MFMO-DE for 2D Darcy Flow dataset.

The architectures with the best, median, and worst RMSE values on the Pareto front (i.e., Point B, Point M, and Point W in Fig. 6) are used for evaluation (named as ACK-MFMO-DE-U-Net-B, ACK-MFMO-DE-U-Net-M, and ACK-MFMO-DE-U-Net-W respectively). Results from two models, i.e., U-Net and Fourier neural operator (FNO), from [48] are used for comparison. The comparison of results using the ACK-MFMO-DE-U-Net, U-Net, and FNO is listed in Table 3. The predictions of different models for 2D Darcy Flow dataset are plotted in Fig. 7.

Table 3 Comparison of results for 2D Darcy Flow dataset.

| Metric | ACK-MFMO-DE-U-Net-B | ACK-MFMO-DE-U-Net-M | ACK-MFMO-DE-U-Net-W | U-Net [48] | FNO [48] |
| --- | --- | --- | --- | --- | --- |
| RMSE | **0.055** | **0.063** | 0.089 | 0.073 | 0.11 |
| nRMSE | 0.0032 | 0.0037 | 0.0053 | 0.0044 | 0.0064 |
| cRMSE | 0.036 | 0.037 | 0.043 | 0.051 | 0.089 |
| bRMSE | 0.051 | 0.097 | 0.226 | 0.046 | 0.079 |
| Max error | **1.4** | 1.9 | 2.0 | 1.7 | 2.1 |
| Number of FLOPs (MB) | 1964 | 1270 | **762** | 2883 | 28.44 |

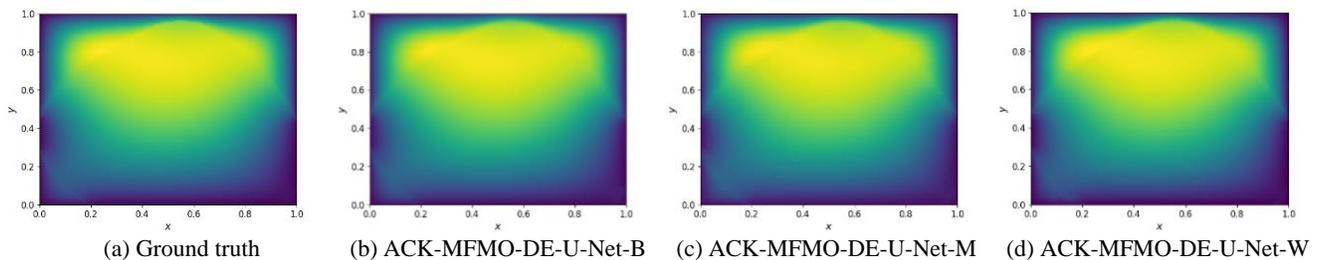

(a) Ground truth　　(b) ACK-MFMO-DE-U-Net-B　　(c) ACK-MFMO-DE-U-Net-M　　(d) ACK-MFMO-DE-U-Net-W

Fig. 7 Plots of predictions from different models for 2D Darcy Flow dataset.

As shown in Table 3, ACK-MFMO-DE-U-Net-B outperforms the U-Net from [48] in terms of RMSE, nRMSE, cRMSE, and max error. Consistent with their placement on the Pareto front obtained from NAS, ACK-MFMO-DE-U-Net-M performs slightly worse than ACK-MFMO-DE-U-Net-B, but it still performs better on RMSE, nRMSE, and cRMSE than the U-Net model from [48]. Similarly, ACK-MFMO-DE-U-Net-W has a poorer predictive performance than ACK-MFMO-DE-U-Net-B and ACK-MFMO-DE-U-Net-M, but has the lowest computational cost in terms of FLOPs. In addition, we note that the number of FLOPs of the models on the Pareto front are lower than that of the U-Net model from [48], despite the significantly better predictive performance, illustrating the importance of employing a NAS algorithm during model training. Though FNO has the least number of FLOPs, the evaluation metrics are significantly worse than other models. Overall, the results illustrate the effectiveness of employing NAS for improving the performance of data-driven regression models for engineering systems such as Darcy Flow.

The downsampling cell structure of ACK-MFMO-DE-U-Net-B is depicted in Fig. 8. It is evident that the optimal architecture found maintains connections between each node and the two preceding cells, enabling comprehensive feature learning by leveraging prior information. This is similar to the ResNet structure, which has been shown to be helpful in prior literature [1]. When the connection to Cell 1 is removed (i.e., sequential connection for downsampling cells akin to basic U-Net), the test RMSE approximately doubles to 0.1, compared to 0.055 for ACK-MFMO-DE-U-Net-B. This demonstrates the effectiveness of the proposed NAS algorithm in finding an optimal architecture from the proposed encoding.

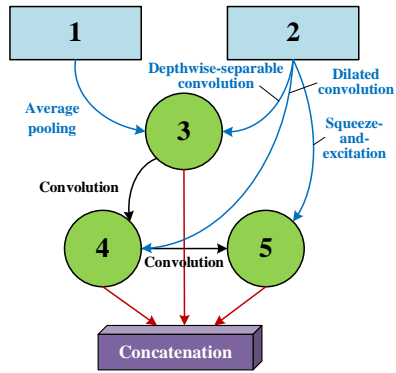

Fig. 8  Structure of the downsampling cell in ACK-MFMO-DE-U-Net-B.

To validate the optimization performance of ACK-MFMO-DE for the NAS problem, the competitive MO-ACK is also used to solve the NAS problem for the same 2D Darcy Flow dataset for three runs. The normalized HV values of the two models are compared in Table 4.

Table 4  Comparison of normalized HV values of ACK-MFMO-DE-U-NET and MO-ACK-U-NET for 2D Darcy Flow dataset.

| Model | Normalized HV value |
| --- | --- |
| ACK-MFMO-DE-U-Net | **0.8418±0.0026** |
| MO-ACK-U-Net | 0.8192±0.0034 |

It can be seen from Table 4 that the normalized HV value of ACK-MFMO-DE-U-Net is significantly larger than that of MO-ACK-U-Net, further indicating better convergence performance of the proposed ACK-MFMO-DE for NAS problems.

5.2.3. NAS Results of CHASE_DB1 Dataset

Similarly, the results for the CHASE_DB1 dataset with median HV value from three randomly-initialized runs are selected for analysis. The Pareto front of different generalized U-Net architectures obtained by ACK-MFMO-DE for CHASE_DB1 dataset is shown in Fig. 9.

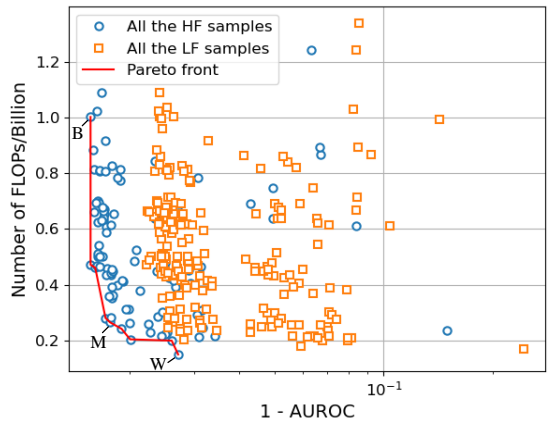

Fig. 9  Pareto front of different generalized U-Net architectures obtained by ACK-MFMO-DE for CHASE_DB1 dataset.

In Fig. 9, ACK-MFMO-DE-U-Net-B, ACK-MFMO-DE-U-Net-M, and ACK-MFMO-DE-U-Net-W also represent the architectures with the best, median, and worst predictive errors (i.e., 1-AUROC) on the Pareto front of NAS. An additional extended training is conducted for these three models to ensure convergence.

To illustrate the effectiveness of the ACK-MFMO-DE-U-Net, several state-of-the-art models are used for comparison, and their corresponding results are extracted from literature and presented in Table 5. The predictions from different models for CHASE_DB1 dataset are plotted in Fig. 10.

In Table 5, the AUROC value of the model with the best predictive performance on the Pareto front (i.e., ACK-MFMO-DE-U-Net-B) outperforms that of multiple state-of-the-art models from literature spanning the last few years, thereby indicating the effectiveness of the NAS algorithm. For ACK-MFMO-DE-U-Net-M and ACK-MFMO-DE-U-Net-W, the AUROC values are slightly worse than ACK-MFMO-DE-U-Net-B while requiring a significantly lower number of FLOPS, as is consistent with their placement on the Pareto front obtained from NAS. The results show that the ACK-MFMO-DE-U-Net effectively identifies a set of high-quality Pareto-optimal architectures that balance the dual objective of predictive performance and computational complexity.

Table 5  Comparison of results for CHASE_DB1 dataset.

| Model | SE | SP | ACC | F1 | AUROC |
|---|---|---|---|---|---|
| MS-NFN [56] | 0.7538 | 0.9847 | 0.9637 | - | 0.9825 |
| Joint-loss framework [57] | 0.7633 | 0.9809 | 0.9610 | - | 0.9781 |
| BTS-DSN [58] | 0.7888 | 0.9801 | 0.9627 | 0.7983 | 0.9840 |
| IterNet [59] | 0.7970 | 0.9823 | 0.9655 | 0.8073 | 0.9851 |
| CSU-Net [60] | 0.8427 | 0.9836 | 0.9706 | 0.8105 | 0.9824 |
| PLRS-Net [61] | 0.8301 | 0.9839 | 0.9731 | - | 0.9863 |
| MRRM-Net [62] | 0.8380 | 0.9841 | 0.9782 | 0.8435 | 0.9887 |
| Genetic U-Net [63] | 0.8463 | 0.9845 | 0.9667 | 0.8223 | 0.9880 |
| DPF-Net [64] | 0.8303 | 0.9839 | 0.9676 | 0.8302 | 0.9868 |
| TP-Net [65] | 0.8600 | 0.9841 | 0.9730 | 0.8518 | 0.9869 |
| AL-ECNAS [66] | 0.8338 | 0.9840 | 0.9743 | 0.8056 | 0.9882 |
| ACK-MFMO-DE-U-Net-B | 0.7663 | 0.9917 | 0.9669 | 0.8362 | **0.9897** |
| ACK-MFMO-DE-U-Net-M | 0.7435 | 0.9930 | 0.9655 | 0.8261 | **0.9893** |
| ACK-MFMO-DE-U-Net-W | 0.7291 | 0.9929 | 0.9638 | 0.8162 | 0.9882 |

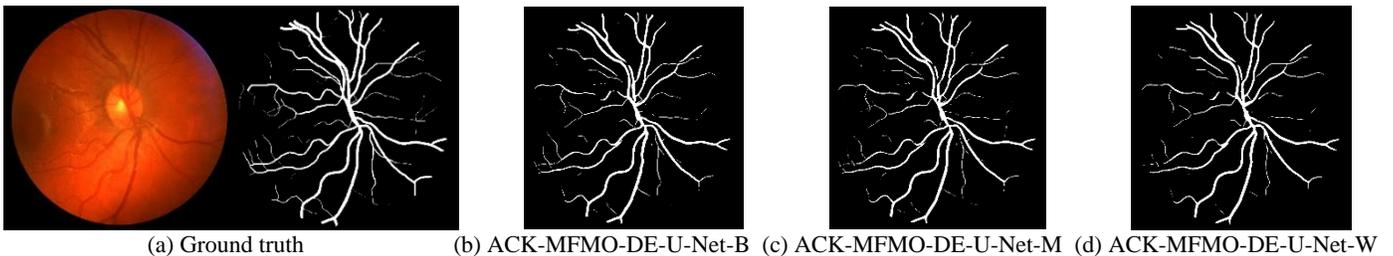

(a) Ground truth        (b) ACK-MFMO-DE-U-Net-B  (c) ACK-MFMO-DE-U-Net-M  (d) ACK-MFMO-DE-U-Net-W
Fig. 10  Plots of predictions of different models for CHASE_DB1 dataset.

Since the Genetic U-Net and AL-ECNAS are also CNNs obtained by evolutionary NAS, the search costs of the three models are compared in Table 6.

Table 6  Comparison of search cost for CHASE_DB1 dataset.

| Model | Genetic U-Net | AL-ECNAS | ACK-MFMO-DE-U-Net |
|---|---|---|---|
| Search cost (GPU-day) | 60 | 4 | **1.467** |

Table 6 shows that the search cost of ACK-MFMO-DE-U-Net is only 2.45% and 36.68% of that of Genetic U-Net and AL-ECNAS, which demonstrates the improved efficiency of the proposed ACK-MFMO-DE-U-Net.

*5.3. Ablation Analysis*

5.3.1. Encoding

To illustrate the effectiveness of the proposed continuous encoding method, the traditional discrete encoding method is also used for NAS. The comparison method is similar to [67]. The NAS with each encoding method is run three times respectively. The normalized HV values of NAS with different encoding methods for 2D Darcy Flow dataset and CHASE_DB1 dataset are compared in Table 7.

Table 7  Comparison of normalized HV values with different encoding methods.

| Dataset | Proposed continuous encoding method | Discrete encoding method |
|---|---|---|
| 2D Darcy Flow | **0.8233±0.0041** | 0.7984±0.0107 |
| CHASE_DB1 | **0.7389±0.0083** | 0.7078±0.0132 |

Table 7 shows that the normalized HV values of the proposed continuous encoding method for NAS are increased by 0.0249 and 0.0311 compared to using the discrete encoding method for 2D Darcy Flow and CHASE_DB1 datasets respectively. This is because the proposed continuous encoding method halves the number of variables, thereby improving the accuracy of the established Co-Kriging model. Simultaneously, the search performance is also improved by customizing the evolution operations with the proposed continuous encoding method and utilizing a tailored infill sampling strategy for surrogate management.

Overall, the results illustrate the effectiveness of the proposed continuous encoding method for NAS.

To illustrate the Co-Kriging model accuracy with iterations, three samples with good RMSE values on Pareto front of 2D Darcy Flow dataset in Fig. 6 are selected for analysis. We discretize the encoding values of Node 5 in Fig. 2 of each sample with step 0.1 in the space. The performances of these discretized points are evaluated through the Co-Kriging model and HF simulation model respectively. The mean squared errors of the predicted values and ground truth for performance are depicted in Fig. 11. It can be seen from Fig. 11 that the Co-Kriging model accuracy in the region around superior samples is significantly improved with iterations, which demonstrates how the proposed encoding method helps improve the accuracy of the Co-Kriging model in the optimization process.

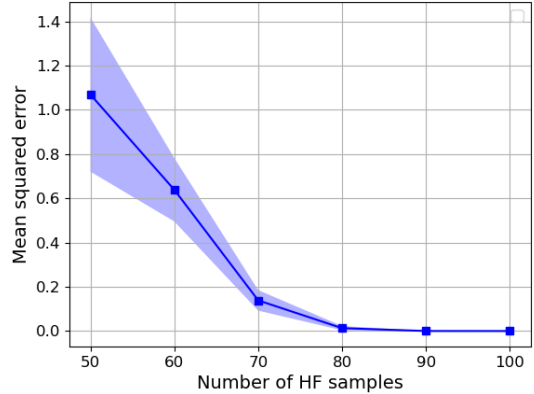

Fig. 11 Mean squared errors of the predicted values and ground truth on performance for 2D Darcy Flow dataset. Shaded area indicates the standard deviation on both sides of the mean.

### 5.3.2. HF Model

To demonstrate the merits of using a multi-fidelity (MF) framework, we also conduct NAS for CHASE_DB1 dataset using only a HF model. To ensure a similar search cost, the $NFE_{max\_HF}$ for a NAS using only a HF model is increased to 150. The normalized HV values of NAS with MF models and HF-only model across three runs are compared in Table 8.

Table 8  Comparison of normalized HV values of NAS with MF models and HF-only model.

| Model | Normalized HV value | Number of samples | Search cost (GPU-day) |
|---|---|---|---|
| NAS with MF models | **0.7798±0.0014** | 100 HF samples and 200 LF samples | 1.467 |
| NAS with HF model | 0.7552±0.0039 | 150 HF samples | 1.675 |

It is evident from Table 8 that the normalized HV value of the NAS with MF models is significantly better than that of the NAS using only HF models even as the search cost of the NAS with MF models is smaller. The results indicate that the use of a MF framework such as the proposed ACK-MFMO-DE-U-Net enhances the ability to find performant architectures under a limited computational budget compared to the use of a single fidelity framework.

### 5.3.3. LF Model

The search performance of our proposed MF model-based NAS is further verified by comparing the NAS with the exclusive use of a LF model to accelerate evaluation. For a fair comparison, the number of function evaluations for the LF-only NAS is increased to 700. Since the predictive performance of NAS results using HF and LF training process is not comparable, all the architectures on the Pareto front of the NAS results obtained using only the LF model are further extended to 25 epochs (i.e., the same number of training epochs as the HF model) to get a consistent set of results for comparison. The median HV value of three NAS runs using MF models is compared to the HV value obtained from the NAS conducted with only the LF model. The normalized HV values for CHASE_DB1 dataset from NAS utilizing both a MF and LF framework are listed in Table 9.

Table 9  Comparison of normalized HV values of NAS with MF models and LF model.

| Model | Normalized HV value | Number of samples | Search cost (GPU-day) |
|---|---|---|---|
| NAS with MF models | **0.7807** | 100 HF samples and 200 LF samples | 1.467 |
| NAS with LF model | 0.7511 | 700 LF samples | 1.485 |

In Table 9, NAS using only the LF model performs worse than when MF models are used. This is because the LF model is fundamentally still different from the true model, preventing the NAS from converging to the real global optimum although it is able to evaluate many more samples in the parameter space. These results further highlight the effectiveness of the proposed MF model-based NAS work.

## 6. Urban Wind Velocity Prediction Problem

In this section, the proposed NAS method is applied to the practical problem of predicting pedestrian-level wind velocities in a complex, real-world urban environment. This problem aims to calculate natural ventilation across different urban building layouts, which can be helpful to urban planners and architects when optimizing estate and precinct design. The dataset is introduced in [68] and was obtained through computational fluid dynamics (CFD) simulations of various real-world urban layouts. The dataset includes 3640 simulations spanning 910 different urban layouts, where each layout is simulated with four wind directions. The simulations for training and validation are further divided into 3240 and 400 samples for this work. Each simulation has three channels (i.e., velocities along $u$, $v$, and $w$ directions), and the layout is represented as a set of 256×256 values. The Adam optimizer, along with a cosine annealing learning rate scheduler and an initial learning rate of 0.001, is used for training, with the batch size set to 16.

Models trained with 50/10 epochs are used as the HF/LF models for architecture search. Mean absolute error (MAE) is adopted as the metric for performance evaluation. After NAS, the Pareto front of different generalized U-Net architectures obtained by ACK-MFMO-DE for this urban problem is depicted in Fig. 12.

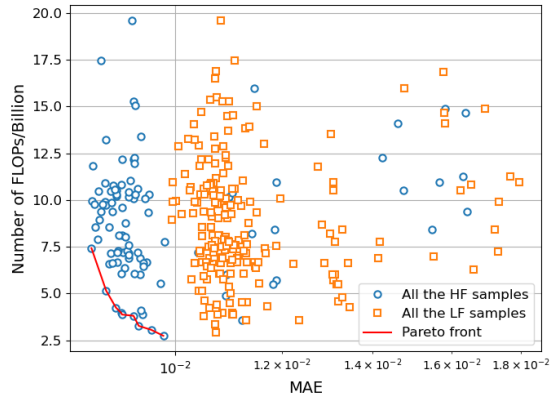

Fig. 12 Pareto front of different generalized U-Net architectures obtained by ACK-MFMO-DE for urban problem.

As per Fig. 12, the proposed ACK-MFMO-DE-U-Net can find a set of architectures with significantly different computational complexity. It indicates that ACK-MFMO-DE-U-Net is capable of finding promising architectures with less computational complexity while ensuring good predictive performance.

The normalized HV values of ACK-MFMO-DE-U-Net and MO-ACK-U-Net for the practical urban problem are compared in Table 10. As per previous results, the normalized HV value of ACK-MFMO-DE-U-Net is increased by about 0.03 relative to MO-ACK-U-Net, which further demonstrates the effectiveness of the proposed NAS method for the practical engineering problem.

Table 10  Comparison of normalized HV values of ACK-MFMO-DE-U-NET and MO-ACK-U-NET for urban problem.

| Model | Normalized HV value |
| --- | --- |
| ACK-MFMO-DE-U-Net | **0.8551** |
| MO-ACK-U-Net | 0.8265 |

## 7. Conclusions

In this paper, an adaptive Co-Kriging-assisted multi-fidelity multi-objective approach is developed to efficiently solve a multi-objective problem that leverages the flexibility and global optimization capability of evolutionary algorithms. We demonstrate this on a set of multi-objective NAS problems where a balance between model predictive performance and computational complexity is required. A novel continuous encoding method for the U-Net-based search space as established by a set of parameterized cells spanning operators and connections is further proposed, which is shown to integrate with and greatly improve the search performance of our proposed multi-fidelity, multi-objective NAS methodology. The effectiveness of the proposed multi-fidelity and multi-objective NAS method has been validated through numerical benchmarks and multiple real-world problems involving 2D Darcy flow regression, CHASE_DB1 biomedical image segmentation, and urban wind velocity prediction, out-performing state-of-the-art results from recent literature. This work thereby shines light on a potential route to alleviating the perennial challenge of huge computational cost of NAS. In future extensions, other types of variable-fidelity models and infill sampling ratio adaptive method for HF/LF samples according to the learned weights balancing HF/LF models in Co-Kriging can be incorporated into the NAS to further improve search performance.

### CRediT authorship contribution statement

**Zhao Wei:** Conceptualization, Methodology, Investigation, Formal analysis, Validation, Visualization, Data curation, Writing

– original draft, Writing – review & editing. **Chin Chun Ooi:** Supervision, Conceptualization, Methodology, Formal analysis, Writing – review & editing. **Yew Soon Ong:** Supervision, Writing – review & editing.

**Declaration of competing interest**

The authors declare that they have no known competing financial interests or personal relationships that could have appeared to influence the work reported in this paper.

**Data availability**

Most of the datasets are publicly available. The dataset for the engineering problem can be provided upon request.

**Acknowledgments**

The financial support of Agency for Science, Technology and Research (A*STAR) is gratefully acknowledged.

**References**

[1] K. He, X. Zhang, S. Ren, and J. Sun, "Deep residual learning for image recognition," in *Proc. IEEE/CVF Conf. Comput. Vis. Pattern Recognit.*, 2016, pp. 770-778.
[2] C. Szegedy, W. Liu, Y. Jia, P. Sermanet, S. Reed, D. Anguelov, D. Erhan, V. Vanhoucke, and A. Rabinovich, "Going deeper with convolutions," in *Proc. IEEE/CVF Conf. Comput. Vis. Pattern Recognit.*, 2015, pp. 1-9.
[3] G. Huang, Z. Liu, L. V. D. Maaten, and K. Q. Weinberger, "Densely connected convolutional networks," in *Proc. IEEE/CVF Conf. Comput. Vis. Pattern Recognit.*, 2017, pp. 4700-4708.
[4] O. Ronneberger, P. Fischer, and T. Brox, "U-Net: Convolutional networks for biomedical image segmentation," in *Med. Image Comput. Computer-Assisted Interv.*, 2015, pp. 234-241.
[5] Z. Lu, R. Cheng, Y. Jin, K. C. Tan, and K. Deb, "Neural architecture search as multiobjective optimization benchmarks: Problem formulation and performance assessment," *IEEE Trans. Evol. Comput.*, vol. 28, no. 2, pp. 323-337, 2024.
[6] J. Huang, B. Xue, Y. Sun, M. Zhang, and G. G. Yen, "Particle swarm optimization for compact neural architecture search for image classification," *IEEE Trans. Evol. Comput.*, vol. 27, no. 5, pp. 1298-1312, 2023.
[7] C. Liu, B. Zoph, M. Neumann, J. Shlens, W. Hua, L. Li, F. Li, A. Yuille, J. Huang, K. Murphy, "Progressive neural architecture search," in *Proc. Eur. Conf. Comput. Vis.*, 2018, pp. 19-34.
[8] T. Elsken, J. H. Metzen, and F. Hutter, "Neural architecture search: A survey," *J. Mach. Learn. Res.*, vol. 20, no. 1, pp. 1997-2017, 2019.
[9] Y. Weng, T. Zhou, Y. Li, and X. Qiu, "NAS-Unet: Neural architecture search for medical image segmentation," *IEEE Access*, vol. 7, pp. 44247-44257, 2019.
[10] B. Zoph and Q. V. Le, "Neural architecture search with reinforcement learning," in *Proc. Int. Conf. Learn. Represent.*, 2016.
[11] H. Liu, K. Simonyan, and Y. Yang, "DARTS: Differentiable architecture search," in *Proc. Int. Conf. Learn. Represent.*, 2018.
[12] Z. Lu, I. Whalen, V. Boddeti, Y. Dhebar, K. Deb, E. Goodman, and W. Banzhaf, "NSGA-Net: Neural architecture search using multi-objective genetic algorithm," in *Proc. Genet. Evol. Comput. Conf.*, 2019, pp. 419-427.
[13] J. Huang, B. Xue, Y. Sun, M. Zhang, and G. G. Yen, "Split-level evolutionary neural architecture search with elite weight inheritance," *IEEE Trans. Neural Networks Learn. Syst.*, pp. 1-15, 2023.
[14] Z. Lu, I. Whalen, Y. Dhebar, K. Deb, E. D. Goodman, W. Banzhaf, and V. N. Boddeti, "Multiobjective evolutionary design of deep convolutional neural networks for image classification," *IEEE Trans. Evol. Comput.*, vol. 25, no. 2, pp. 277-291, 2021.
[15] R. Vinuesa, H. Azizpour, I. Leite, M. Balaam, V. Dignum, S. Domisch, A. Felländer, S. D. Langhans, M. Tegmark, and F. F. Nerini, "The role of artificial intelligence in achieving the Sustainable Development Goals," *Nat. Commun.*, vol. 11, no. 233, pp. 1-10, 2020.
[16] Y. Liu, Y. Sun, B. Xue, M. Zhang, G. G. Yen, and K. C. Tan, "A survey on evolutionary neural architecture search," *IEEE Trans. Neural Networks Learn. Syst.*, vol. 34, no. 2, pp. 550-570, 2023.
[17] L. Wen, L. Gao, X. Li, and H. Li, "A new genetic algorithm based evolutionary neural architecture search for image classification," *Swarm Evol. Comput.*, vol. 75, pp. 101191, 2022.
[18] X. Chu, B. Zhang, and R. Xu, "Multi-objective reinforced evolution in mobile neural architecture search," in *Proc. Eur. Conf. Comput. Vis.*, 2020, pp. 99-113.
[19] Y. Jin, "Surrogate-assisted evolutionary computation: Recent advances and future challenges," *Swarm Evol. Comput.*, vol. 1, no. 2, pp. 61-70, 2011.
[20] D. Lim, Y. Jin, Y. S. Ong, and B. Sendhoff, "Generalizing surrogate-assisted evolutionary computation," *IEEE Trans. Evol. Comput.*, vol. 14, no. 3, pp. 329-355, 2010.
[21] J. F. Gonçalves and M. G. C. Resende, "Biased random-key genetic algorithms for combinatorial optimization," *J. Heuristics*, vol. 17, no. 5, pp. 487-525, 2011.
[22] L. V. Snyder and M. S. Daskin, "A random-key genetic algorithm for the generalized traveling salesman problem," *Eur. J. Oper. Res.*, vol. 174, no. 1, pp. 38-53, 2006.
[23] K. T. Chitty-Venkata, M. Emani, V. Vishwanath, and A. K. Somani, "Neural architecture search benchmarks: Insights and survey," *IEEE Access*, vol. 11, pp. 25217-25236, 2023.
[24] Z. Zhou, M. M. R. Siddiquee, N. Tajbakhsh, and J. Liang, "Unet++: A nested U-Net architecture for medical image segmentation," *Deep Learn. Med. Image Anal. Multimodal Learn. Clin. Decis. Support*, vol. 11045, pp. 3-11, 2018.
[25] O. Oktay, J. Schlemper, L. L. Folgoc, M. Lee, M. Heinrich, K. Misawa, K. Mori, S. McDonagh, N. Y. Hammerla, B. Kainz, B. Glocker, and D. Rueckert, "Attention U-Net: Learning where to look for the pancreas," in *Conf. Med. Imaging Deep Learn.*, 2018.
[26] E. Real, S. Moore, A. Selle, S. Saxena, Y. L. Suematsu, J. Tan, Q. V. Le, and A. Kurakin, "Large-scale evolution of image classifiers," in *Proc. Int. Conf. Mach. Learn.*, 2017, pp. 2902-2911.

[27] Y. Xue, C. Chen, and A. Słowik, "Neural architecture search based on a multi-objective evolutionary algorithm with probability stack," *IEEE Trans. Evol. Comput.*, vol. 27, no. 4, pp. 778-786, 2023.

[28] T. Zhang, C. Lei, Z. Zhang, X. B. Meng, and C. P. Chen, "AS-NAS: Adaptive scalable neural architecture search with reinforced evolutionary algorithm for deep learning," *IEEE Trans. Evol. Comput.*, vol. 25, no. 5, pp. 830-841, 2021.

[29] Y. Tian, S. Peng, S. Yang, X. Zhang, K. C. Tan, and Y. Jin, "Action command encoding for surrogate-assisted neural architecture search," *IEEE Trans. Cogn. Commun. Netw.*, vol. 14, no. 3, pp. 1129-1142, 2022.

[30] L. Fan and H. Wang, "Surrogate-assisted evolutionary neural architecture search with network embedding," *Complex Intell. Syst.*, vol. 9, no. 3, pp. 3313-3331, 2023.

[31] Y. Zhou, Y. Jin, Y. Sun, and J. Ding, "Surrogate-assisted cooperative co-evolutionary reservoir architecture search for liquid state machines," *IEEE Tran. Emerg. Top. Comput. Intell.*, vol. 7, no. 5, pp. 1484-1498, 2023.

[32] Y. Liu and J. Liu, "A surrogate evolutionary neural architecture search algorithm for graph neural networks," *Appl. Soft Comput.*, vol. 144, pp. 110485, 2023.

[33] K. Lyu, H. Li, M. Gong, L. Xing, and A. K. Qin, "Surrogate-assisted evolutionary multiobjective neural architecture search based on transfer stacking and knowledge distillation," *IEEE Trans. Evol. Comput.*, vol. 28, no. 3, pp. 608-622, 2024.

[34] Z. Lu, R. Cheng, S. Huang, H. Zhang, C. Qiu, and F. Yang, "Surrogate-assisted multiobjective neural architecture search for real-time semantic segmentation," *IEEE Trans. Artif. Intell.*, vol. 4, no. 6, pp. 1602-1615, 2023.

[35] Z. Lu, K. Deb, E. Goodman, W. Banzhaf, and V. N. Boddeti, "NSGANetV2: Evolutionary multi-objective surrogate-assisted neural architecture search," in *Proc. Eur. Conf. Comput. Vis.*, 2020, pp. 35-51.

[36] L. Zimmer, M. Lindauer, and F. Hutter, "Auto-Pytorch: Multi-fidelity MetaLearning for efficient and robust AutoDL," *IEEE Trans. Pattern Anal. Mach. Intell.*, vol. 43, no. 9, pp. 3079-3090, 2021.

[37] I. Trofimov, N. Klyuchnikov, M. Salnikov, A. Filippov, and E. Burnaev, "Multi-fidelity neural architecture search with knowledge distillation," *IEEE Access*, vol. 11, pp. 59217-59225, 2023.

[38] J. Xu, N. A. V. Suryanarayanan, and H. Iba, "MPENAS: Multi-fidelity predictor-guided evolutionary neural architecture search with zero-cost proxies," *Proc. Genet. Evol. Comput. Conf.*, 2023, pp. 1276-1285.

[39] S. Yang, Y. Tian, X. Xiang, S. Peng, and X. Zhang, "Accelerating evolutionary neural architecture search via multifidelity evaluation," *IEEE Trans. Cogn. Dev. Syst.*, vol. 14, no. 4, pp. 1778-1792, 2022.

[40] B. Zoph, V. Vasudevan, J. Shlens, and Q. V. Le, "Learning transferable architectures for scalable image recognition," in *Proc. IEEE/CVF Conf. Comput. Vis. Pattern Recognit.*, 2018, pp. 8697-8710.

[41] L. Xie and A. Yuille, "Genetic CNN," in *Proc. Int. Conf. Comput. Vis.*, 2017, pp. 1379-1388.

[42] Y. Sun, B. Xue, M. Zhang, and G. G. Yen, "Evolving deep convolutional neural networks for image classification," *IEEE Trans. Evol. Comput.*, vol. 24, no. 2, pp. 394-407, 2020.

[43] Z. Yang, Y. Wang, X. Chen, B. Shi, C. Xu, C. Xu, Q. Tian, and C. Xu "CARS: Continuous evolution for efficient neural architecture search," in *Proc. IEEE/CVF Conf. Comput. Vis. Pattern Recognit.*, 2020, pp. 1829-1838.

[44] Y. Wang, L. Zhen, J. Zhang, M. Li, L. Zhang, Z. Wang, Y. Feng, Y. Xue, X. Wang, Z. Chen, T. Luo, R. S. M. Goh, and Y. Liu, "MedNAS: Multi-scale training-free neural architecture search for medical image analysis," *IEEE Trans. Evol. Comput.*, vol. 28, no. 3, pp. 668-681, 2024.

[45] L. Ma, N. Li, G. Yu, X. Geng, S. Cheng, X. Wang, M. Huang, and Y. Jin, "Pareto-wise ranking classifier for multi-objective evolutionary neural architecture search," *IEEE Trans. Evol. Comput.*, vol. 28, no. 3, pp. 570-581, 2024.

[46] A. I. Forrester, A. Sóbester, and A. J. Keane, "Multi-fidelity optimization via surrogate modelling," *Proc. R. Soc. A: Math. Phys. Eng. Sci.*, vol. 463, pp. 3251-3269, 2007.

[47] K. Deb, A. Pratap, S. Agarwal, and T. Meyarivan, "A fast and elitist multiobjective genetic algorithm: NSGA-II," *IEEE Trans. Evol. Comput.*, vol. 6, no. 2, pp. 182-197, 2002.

[48] M. Takamoto, T. Praditia, R. Leiteritz, D. MacKinlay, F. Alesiani, D. Pflüger, and M. Niepert, "PDEBench: An extensive benchmark for scientific machine learning," in *Proc. Adv. Neural Inf. Process. Syst.*, vol. 35, pp. 1596-1611, 2022.

[49] C. G. Owen, A. R. Rudnicka, R. Mullen, S. A. Barman, D. Monekosso, P. H. Whincup, J. Ng, and C. Paterson, "Measuring retinal vessel tortuosity in 10-year-old children: Validation of the computer-assisted image analysis of the retina (CAIAR) program," *Invest. Ophthalmol. Vis. Sci.*, vol. 50, no. 5, pp. 2004-2010, 2009.

[50] R. Shi, T. Long, and H. Baoyin, "Multi-fidelity and multi-objective optimization of low-thrust transfers with control strategy for all-electric geostationary satellites," *Acta Astronaut.*, vol. 177, pp. 577-587, 2020.

[51] Z. Li, K. Tian, S. Zhang, and B. Wang, "Efficient multi-objective CMA-ES algorithm assisted by knowledge-extraction-based variable-fidelity surrogate model," *Chinese J. Aeronaut.*, vol. 36, no. 6, pp. 213-232, 2023.

[52] L. Shu, P. Jiang, Q. Zhou, and T. Xie, "An online variable-fidelity optimization approach for multi-objective design optimization," *Struct. Multidisc. Optim.*, vol. 60, pp. 1059-1077, 2019.

[53] H. Hao, A. Zhou, H. Qian, and H. Zhang, "Expensive multiobjective optimization by relation learning and prediction," *IEEE Trans. Evol. Comput.*, vol. 26, no. 5, pp. 1157-1170, 2022.

[54] Z. Liu and H. Wang, "A data augmentation based Kriging-assisted reference vector guided evolutionary algorithm for expensive dynamic multi-objective optimization," *Swarm Evol. Comput.*, vol. 75, p. 101173, 2022.

[55] A. Habib, H. K. Singh, T. Chugh, T. Ray, and K. Miettinen, "A multiple surrogate assisted decomposition-based evolutionary algorithm for expensive multi/many-objective optimization," *IEEE Trans. Evol. Comput.*, vol. 23, no. 6, pp. 1000-1014, 2019.

[56] Y. Wu, Y. Xia, Y. Song, Y. Zhang, and W. Cai, "Multiscale network followed network model for retinal vessel segmentation," in *Med. Image Comput. Computer-Assisted Interv.*, 2018, pp. 119-126.

[57] Z. Yan, X. Yang, and K. T. Cheng, "Joint segment-level and pixel-wise losses for deep learning based retinal vessel segmentation," *IEEE Trans. Biomed. Eng.*, vol. 65, no. 9, pp. 1912-1923, 2018.

[58] S. Guo, K. Wang, H. Kang, Y. Zhang, Y. Gao, and T. Li, "BTS-DSN: Deeply supervised neural network with short connections for retinal vessel segmentation," *Int. J. Med. Inform.*, vol. 126, pp. 105-113, 2019.

[59] L. Li, M. Verma, Y. Nakashima, H. Nagahara, and R. Kawasaki, "IterNet: Retinal image segmentation utilizing structural redundancy in

vessel networks," in *Proc. IEEE/CVF Winter Conf. Appl. Comput. Vis.*, 2020, pp. 3656-3665.

[60] B. Wang, S. Wang, S. Qiu, W. Wei, H. Wang, and H. He, "CSU-Net: A context spatial U-Net for accurate blood vessel segmentation in fundus images," *IEEE J. Biomed. Health Inform.*, vol. 25, no. 4, pp. 1128-1138, 2021.

[61] M. Arsalan, A. Haider, Y. W. Lee, and K. R. Park, "Detecting retinal vasculature as a key biomarker for deep Learning-based intelligent screening and analysis of diabetic and hypertensive retinopathy," *Expert Syst. Appl.*, vol. 200, pp. 117009, 2022.

[62] J. Pan, J. Gong, M. Yu, J. Zhang, Y. Guo, and G. Zhang, "A multilevel remote relational modeling network for accurate segmentation of fundus blood vessels," *IEEE Trans. Instrum. Meas.*, vol. 71, pp. 1-14, 2022.

[63] J. Wei, G. Zhu, Z. Fan, J. Liu, Y. Rong, J. Mo, W. Li, and X. Chen, "Genetic U-Net: Automatically designed deep networks for retinal vessel segmentation using a genetic algorithm," *IEEE Trans. Med. Imaging*, vol. 41, no. 2, pp. 292-307, 2022.

[64] J. Li, G. Gao, L. Yang, G. Bian, and Y. Liu, "DPF-Net: A dual-path progressive fusion network for retinal vessel segmentation," *IEEE Trans. Instrum. Meas.*, vol. 72, pp. 1-17, 2023.

[65] Z. Qu, L. Zhuo, J. Cao, X. Li, H. Yin, and Z. Wang, "TP-Net: Two-path network for retinal vessel segmentation," *IEEE J. Biomed. Health Inform.*, vol. 27, no. 4, pp. 1979-1990, 2023.

[66] W. Wang, X. Wang, and X. Song, "Adaptive lightweight convolutional neural architecture search for segmentation problem," *Eng. Optimiz.*, pp. 1-18, 2023.

[67] Y. Wang, H. Liu, H. Long, Z. Zhang, and S. Yang, "Differential evolution with a new encoding mechanism for optimizing wind farm layout," *IEEE Trans. Industr. Inform.*, vol. 14, no. 3, pp. 1040-1054, 2018.

[68] S. J. Low, V. S. G. Raghavan, H. Gopalan, J. C. Wong, J. Yeoh, and C. C. Ooi, "FastFlow: AI for fast urban wind velocity prediction," in *IEEE Int. Conf. Data Min. Workshops*, 2022, pp. 147-154.